\documentclass{article}
\usepackage[english]{babel}

\usepackage[letterpaper,top=2cm,bottom=2cm,left=3cm,right=3cm,marginparwidth=1.75cm]{geometry}

\usepackage{amsmath}
\usepackage{graphicx}
\usepackage[colorlinks=true, allcolors=blue]{hyperref}
\usepackage[affil-it]{authblk} % Load the authblk package
\usepackage{tikz}
\usepackage{multirow}
\usepackage{subcaption}

\title{Quantifying Distribution Shifts and Uncertainties for Enhanced Model Robustness in Machine Learning Applications}

\author{Vegard Flovik \footnote{\texttt{vflovik@gmail.com}}}
\affil{Department of Electronic Systems, Norwegian University of Science and Technology, Trondheim, Norway} % Author affiliation

\begin{document}
\maketitle

\begin{abstract}
Distribution shifts, where statistical properties differ between training and test datasets, present a significant challenge in real-world machine learning applications where they directly impact model generalization and robustness. 
In this study, we explore model adaptation and generalization by utilizing synthetic data to systematically address distributional disparities. Our investigation aims to identify the prerequisites for successful model adaptation across diverse data distributions, while quantifying the associated uncertainties. Specifically, we generate synthetic data using the van der Waals equation for gases and employ quantitative measures such as Kullback-Leibler divergence, Jensen-Shannon distance, and Mahalanobis distance to assess data similarity. These metrics enable us to evaluate both model accuracy and quantify the associated uncertainty in predictions arising from data distribution shifts. 
Our findings suggest that utilizing statistical measures, such as the Mahalanobis distance, to determine whether model predictions fall within the low-error "interpolation regime" or the high-error "extrapolation regime" provides a complementary method for assessing distribution shift and model uncertainty. These insights hold significant value for enhancing model robustness and generalization, essential for the successful deployment of machine learning applications in real-world scenarios.

\end{abstract}

\section{Introduction}

In machine learning, ensuring model accuracy and reliability is paramount for successful deployment in real-world applications. However, challenges arise when the statistical properties of the data differ between training and test datasets, a phenomenon commonly referred to as distribution shift \cite{quinonero2022dataset}. 
This presents significant challenges in dynamic systems where data distributions evolve over time, as well as in transfer learning scenarios where models encounter disparate data distributions \cite{pan2010survey}. Despite efforts to align source and target domains, practical constraints often lead to distributional disparities, impacting model accuracy and reliability in real-world applications \cite{farahani2021brief}.

Numerous examples illustrate the implications of distribution shift for machine learning models across various domains.
Consider, for example, a scenario where machine learning models monitor the performance of wind turbines in a wind farm. Covariate shift occurs if environmental factors affecting turbine performance, such as wind speed, wind direction, and temperature, differ between historical training data and current operational data due to changes in weather patterns or site conditions. Conversely, target drift arises if the relationship between these environmental factors and turbine performance changes over time due to factors like aging equipment or modifications in operational protocols.

Similarly, in medical image analysis, models trained to detect tumors or diseases from X-rays may fail when deployed in different hospitals with varying equipment, leading to potential misdiagnoses and harmful outcomes for patients \cite{ellis2022twelve}. Autonomous vehicles may encounter unforeseen scenarios not adequately represented in training data, posing substantial risks to passenger safety and public trust in autonomous driving technology \cite{muhammad2020deep}. Recommendation systems deployed in online platforms may provide biased or inaccurate suggestions when user preferences evolve over time or when operating in new contexts, leading to sub-optimal user experiences and potential ethical concerns \cite{rabiu2020recommender}.
These examples underscore the critical importance of addressing distribution shift and its impact on model performance in real-world applications spanning numerous industries and domains.

Previous studies by Shimodaira \cite{shimodaira_improving} and Liu et al. \cite{liu2021towards} have also shed light on the detrimental effects of distributional disparities between training and test datasets on model performance and generalization. These findings have initiated the exploration of various domain adaptation methods aimed at mitigating these disparities. For example, adversarial training \cite{ganin2016domain} and discrepancy-based approaches \cite{long2015learning}, have emerged as potential solutions to align source and target domain, thereby reducing distributional disparities.  
Furthermore, data augmentation techniques, including mixup \cite{zhang2017mixup} and generative adversarial networks (GANs) \cite{goodfellow2014generative}, have also been explored to enhance model robustness against distribution shift.

Additionally, Bayesian methods \cite{gawlikowski2023survey} and conformal prediction \cite{conformal_prediction} offer alternative approaches for uncertainty quantification in machine learning. Bayesian methods provide a probabilistic framework for uncertainty estimation, modeling uncertainty as a distribution over parameters or predictions. Conformal prediction, on the other hand, offers a pragmatic approach by providing prediction intervals with valid coverage guarantees. Integrating these methods into machine learning pipelines can provide additional insights into model uncertainty and enhance model robustness in real-world scenarios.

Previous work by Ovadia et al. \cite{NEURIPS2019_8558cb40} have also emphasized the importance of quantifying distribution shift and its impact on model uncertainty. Proper quantitative measures provide valuable insights into the extent of distributional disparities, and these measures serve as critical tools for assessing the robustness and generalization abilities of machine learning models.

In this study, we aim to address a gap in understanding distribution shift and its impact on model performance by investigating the role of synthetic data in evaluating model robustness. Leveraging synthetic data generated using the Van der Waals equation for gases, we systematically quantify distributional disparities and assess their impact on model performance and uncertainty.

\subsection{Research Questions}
This study aims to answer the following fundamental questions surrounding the challenges with data distribution shifts:

\begin{itemize}
  \item How can we quantify the degree of distribution shift and data similarity?
  \item How well can machine learning models generalize across different data sets?
  \item What uncertainties arise from differences in data distribution between train and test data, and how can we quantify this?
\end{itemize}
By systematically investigating these questions through a comprehensive analysis of data similarity measures and model uncertainty metrics, we aim to advance our understanding of distribution shift and provide actionable insights for enhancing the robustness and generalization of machine learning models in real-world applications.

\subsection*{Paper Structure}
The paper is structured as follows: 

\begin{itemize}
    \item Section \ref{sec:objectives} provides an overview of the main objectives and outlines the methodology employed in the study.
    \item Section \ref{sec:data_similarity} introduces the various measures and metrics used for quantifying data similarity and their application within our study.
    \item Section \ref{sec:Model} introduces the architecture of our model and details the Monte Carlo Dropout method for assessing model uncertainty. 
    \item Section \ref{sec:data_generation_van_der_waals} describes the process of generating our synthetic dataset using the van der Waals equation.
    \item Section \ref{sec:Results} presents the technical details and results of our analysis.
    \item Section \ref{sec:summary} concludes the paper by summarizing the main outcomes and discussing the conclusions drawn from our experiments.
\end{itemize}

\section{Objectives and Methodology}
\label{sec:objectives}
To investigate these research questions in further detail, we perform two different experiments, both based on generating synthetic data where we have full control of the distributional properties across the different datasets. Further details and methodology, including data generation and analysis, is described in the following sections for experiments 1 and 2 respectively. 

\subsection{Experiment 1: Changes in Feature-Target Correlations}
\textbf{Objective}:
Experiment 1 aims to explore how variations in feature-target correlations impact model accuracy and how data similarity can be quantified. By systematically varying the feature-target correlations across different datasets, we seek insights into the sensitivity of model performance to these variations.
\\
\\
\textbf{Significance}:
Understanding the effect of feature-target correlations and distribution shift on model accuracy is crucial for developing robust models and for assessing their ability to generalize to unseen data beyond the training set.
\\
\\
\textbf{Methodology}:
To investigate this, we employ the "ideal gas" approximation to generate synthetic training data. Subsequently, we generate datasets for different gases, each with properties that deviate from ideal gas behavior, approximated by the Van der Waals equation. In addition to visual comparisons of the data distributions, we explore measures such as the Kullback-Leibler Divergence and Jensen-Shannon Distance to quantify data similarity. Subsequently, we train a machine learning model using the ideal gas data and proceed to predict the properties of the other gases, assessing how distribution shift impacts model accuracy.

\subsection{Experiment 2: Feature Distribution Drift}

\textbf{Objective}:
Experiment 2 aims to investigate changes in the feature distribution and their impact on model accuracy and uncertainty, with a focus on quantifying these properties.
\\
\\
\textbf{Significance}:
In real-world dynamical systems, variations in pressure, temperature, or other properties over time can induce shifts in the feature distribution, referred to as covariate shift. Detecting when the model operates beyond the bounds of its training distribution and understanding the implications for prediction accuracy and uncertainty are thus crucial for maintaining model reliability in practical applications.
\\
\\
\textbf{Methodology}:
We simulate changes in the feature distribution between the training and test datasets using the ideal gas approximation. For each data point in the test set, we measure the deviation of its feature values from the training data distribution using the Mahalanobis distance. Subsequently, we analyze model predictions and evaluate how distribution shift influences both accuracy and uncertainty. During inference, we employ Monte Carlo Dropout as an estimate of the model's uncertainty, which is then correlated with the degree of distribution shift quantified through the Mahalanobis distance.

\section{Quantifying Data Similarity}
\label{sec:data_similarity}

A key aspect of our study involves quantifying the similarity between the various datasets generated by the van der Waals equation. 
In this section, we provide a brief theoretical background on the key metrics used to quantify data similarity and distributional shifts, namely KL-Divergence, Jensen-Shannon Distance, and Mahalanobis Distance. 
These metrics allow us to quantify the extent to which one dataset diverges from another. We can thus evaluate the overall difference or similarity of the various data distributions, and how this affects the model's ability to generalize beyond the training data.

\subsection{Kullback-Leibler Divergence (KL-Divergence)}

KL-Divergence serves as a measure of relative entropy between two probability distributions \cite{kl_divergence}. Given two probability distributions $P(x)$ and $Q(x)$, KL-Divergence is defined as:

\begin{equation}
D_{\text{KL}}(P \| Q) = \sum_{x \in X} P(x) \log \frac{P(x)}{Q(x)}
\end{equation}
KL-Divergence quantifies the information lost when $Q(x)$ is used to approximate $P(x)$. It is non-negative and equals zero if and only if the distributions are identical.

In machine learning and statistics, it has been widely used in tasks such as domain adaptation, transfer learning, and anomaly detection to quantify the dissimilarity between data distributions \cite{kl_divergence_ML}. In the context of comparing data distribution shifts in this study, KL-Divergence thus provides a quantitative measure of how much one dataset, $P(x)$, differs from another, $Q(x)$.

\subsection{Jensen-Shannon Distance}

The Jensen-Shannon Distance is a symmetric measure of the similarity between two probability distributions. It is derived from the KL-Divergence and is defined as:

\begin{equation}
D_{\text{JS}}(P, Q) = \frac{1}{2} D_{\text{KL}}(P \| M) + \frac{1}{2} D_{\text{KL}}(Q \| M)
\end{equation}
where $M$ is the average distribution of $P$ and $Q$.

Unlike KL-Divergence, which is a divergence measure, Jensen-Shannon Distance is a proper metric. It combines elements of KL-Divergence to evaluate the overall difference or similarity between datasets in a more balanced manner. Furthermore, it is bounded between 0 and 1, where 0 indicates identical distributions and 1 indicates completely different distributions. This bounded nature of the Jensen-Shannon Distance provides some advantages when used as a metric to quantify data distribution similarity. In the machine learning literature, Jensen-Shannon Distance is often employed in tasks such as clustering, generative modeling, and classification to assess the similarity between probability distributions arising from different datasets or model predictions \cite{jensen_shannon_distance}.

\subsection{Mahalanobis Distance}
\label{sec:mahalanobis}

In statistical analysis and pattern recognition, the Mahalanobis distance is a metric used to measure the distance between a point and a distribution \cite{mahalanobis_distance}. Unlike Euclidean distance, which treats all dimensions equally, the Mahalanobis distance adjusts for the covariance structure of the data, providing a more accurate measure of similarity or dissimilarity.

Mathematically, the Mahalanobis distance $D_M(x)$ for a data point $x$ from a distribution with mean vector $\mu$ and covariance matrix $S$ is defined as:

\begin{equation}
D_M(x) = \sqrt{ (x - \mu)^T S^{-1} (x - \mu) }
\end{equation}

Here:
\begin{itemize}
  \item $x$ is the vector representing the data point.
  \item $\mu$ is the mean vector of the distribution.
  \item $S$ is the covariance matrix of the distribution.
\end{itemize}
In this study, we utilize the Mahalanobis distance to quantify the similarity of data points in the test set from the training distribution. 
By assessing whether model predictions fall within the known data distribution (interpolation) or extend beyond it (extrapolation), we obtain a deeper understanding of the model's ability to generalize and extrapolate beyond the training data. 

\section{Model Architecture and Uncertainty Quantification}
\label{sec:Model}
The selected model architecture in this study is deliberately kept simple, allowing us to focus on the core research objectives related to data distribution shifts and uncertainties rather than complex modeling approaches. The model architecture includes the following components:

\begin{enumerate}
    \item \textbf{Input Layer}: This layer defines the shape of the input data.
    
    \item \textbf{Dense Layers}: The model includes three dense layers with 64, 64, and 32 neurons respectively. Each neuron in a dense layer is connected to every neuron in the previous layer. The activation function used in these layers is chosen as the Exponential Linear Unit (ELU). 
    
    \item \textbf{Custom Dropout Layer}: This model incorporates a custom dropout layer with a dropout rate of 0.1, randomly setting 10\% of input units to zero. The custom layer also enables dropout during inference to facilitate the Monte Carlo Dropout technique, as discussed in Section \ref{sec:monte_carlo_dropout}.

    \item \textbf{Output Layer}: The final layer consists of a single neuron, representing the output of the model. It does not apply any activation function, which is standard for regression models where the goal is to predict continuous numerical values.

\end{enumerate}
For additional details about the technical implementation, readers are directed to the code available on GitHub: \url{https://github.com/veflo/uncert_quant}

\subsection{Assessing Model Uncertainty with Monte Carlo Dropout}
\label{sec:monte_carlo_dropout}

In this study, we employ Monte Carlo Dropout as a technique to estimate model uncertainty during inference. This approach is particularly relevant for capturing epistemic uncertainty arising from the model's lack of knowledge or understanding about certain regions of the input space \cite{gal2016dropout}.

Dropout is a regularization technique commonly used during model training. It works by randomly "dropping out" (i.e., setting to zero) a fraction of the units in a layer during each training iteration. This prevents units from co-adapting too much and ensures that the network learns a more robust and generalizable representation of the data.
Typically, dropout is turned off during inference to obtain deterministic predictions. However, in Monte Carlo Dropout, dropout is kept active during inference, and multiple forward passes through the network are performed for the same input, resulting in different predictions $y_i$ sampled from the model's predictive distribution:

\begin{equation}
y_i \sim p(y | x, \theta) 
\end{equation}
where $\theta$ represents the model parameters.

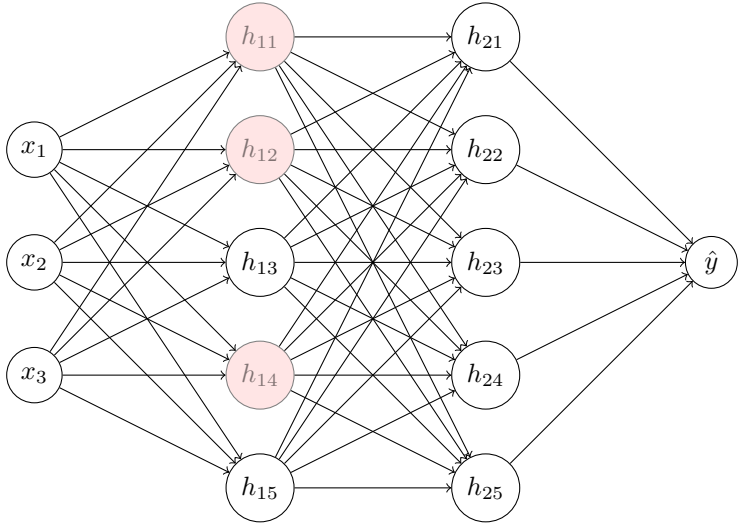
\begin{figure}[htbp]
    \centering
    \begin{tikzpicture}[x=1.5cm, y=1.5cm]
        \node[circle, draw] (x1) at (0,0) {$x_1$};
        \node[circle, draw] (x2) at (0,-1) {$x_2$};
        \node[circle, draw] (x3) at (0,-2) {$x_3$};
        
        % Hidden layer 1
        \node[circle, draw, fill=red!20, opacity=0.5] (h11) at (2,1) {$h_{11}$};
        \node[circle, draw, fill=red!20, opacity=0.5] (h12) at (2,0) {$h_{12}$};
        \node[circle, draw] (h13) at (2,-1) {$h_{13}$};
        \node[circle, draw, fill=red!20, opacity=0.5] (h14) at (2,-2) {$h_{14}$};
        \node[circle, draw] (h15) at (2,-3) {$h_{15}$};
        
        % Hidden layer 2
        \node[circle, draw] (h21) at (4,1) {$h_{21}$};
        \node[circle, draw] (h22) at (4,0) {$h_{22}$};
        \node[circle, draw] (h23) at (4,-1) {$h_{23}$};
        \node[circle, draw] (h24) at (4,-2) {$h_{24}$};
        \node[circle, draw] (h25) at (4,-3) {$h_{25}$};
        
        \node[circle, draw] (y) at (6,-1) {$\hat{y}$};
        
        % Connections
        \draw[->] (x1) -- (h11);
        \draw[->] (x1) -- (h12);
        \draw[->] (x1) -- (h13);
        \draw[->] (x1) -- (h14);
        \draw[->] (x1) -- (h15);
        
        \draw[->] (x2) -- (h11);
        \draw[->] (x2) -- (h12);
        \draw[->] (x2) -- (h13);
        \draw[->] (x2) -- (h14);
        \draw[->] (x2) -- (h15);
        
        \draw[->] (x3) -- (h11);
        \draw[->] (x3) -- (h12);
        \draw[->] (x3) -- (h13);
        \draw[->] (x3) -- (h14);
        \draw[->] (x3) -- (h15);
        
        \draw[->] (h11) -- (h21);
        \draw[->] (h11) -- (h22);
        \draw[->] (h11) -- (h23);
        \draw[->] (h11) -- (h24);
        \draw[->] (h11) -- (h25);
        
        \draw[->] (h12) -- (h21);
        \draw[->] (h12) -- (h22);
        \draw[->] (h12) -- (h23);
        \draw[->] (h12) -- (h24);
        \draw[->] (h12) -- (h25);
        
        \draw[->] (h13) -- (h21);
        \draw[->] (h13) -- (h22);
        \draw[->] (h13) -- (h23);
        \draw[->] (h13) -- (h24);
        \draw[->] (h13) -- (h25);
        
        \draw[->] (h14) -- (h21);
        \draw[->] (h14) -- (h22);
        \draw[->] (h14) -- (h23);
        \draw[->] (h14) -- (h24);
        \draw[->] (h14) -- (h25);
        
        \draw[->] (h15) -- (h21);
        \draw[->] (h15) -- (h22);
        \draw[->] (h15) -- (h23);
        \draw[->] (h15) -- (h24);
        \draw[->] (h15) -- (h25);
        
        \draw[->] (h21) -- (y);
        \draw[->] (h22) -- (y);
        \draw[->] (h23) -- (y);
        \draw[->] (h24) -- (y);
        \draw[->] (h25) -- (y);
        
    \end{tikzpicture}
    \caption{Neural Network with Dropout, as indicated by red nodes}
\end{figure}
We can then aggregate this ensemble of predictions and calculate the mean, $\mu$, and standard deviation, $\sigma$, as follows:

\begin{equation}
\mu = \frac{1}{N} \sum_{i=1}^{N} y_i 
\end{equation}

\begin{equation}
\sigma = \sqrt{\frac{1}{N} \sum_{i=1}^{N} (y_i - \mu)^2} 
\end{equation}
where $N$ is the number of forward passes.

 By performing Monte Carlo Dropout during inference and investigating the correlation between model uncertainty and the degree of distribution shift, as quantified through the Mahalanobis distance, we thus aim to gain insights into the model's ability to generalize to unseen data with different statistical properties.

\section{Data Generation using the van der Waals Equation}
\label{sec:data_generation_van_der_waals}

The van der Waals equation is an equation of state for gases that provides a more accurate description of the behavior of real gases, especially at high pressures and low temperatures, where the assumptions of the ideal gas law start to break down \cite{vdw_equation}.

To generate synthetic datasets for our experiments, we utilize this equation to simulate the behavior of gases under various conditions.
Specifically, we employ the equation to generate datasets for different gases, each exhibiting deviations from ideal gas behavior, thus creating a range of feature-target correlations.

The van der Waals equation is given by:

\begin{equation}
\left( P + \frac{aN^2}{V^2} \right) (V - Nb) = NRT
\end{equation}

Where:
\begin{itemize}
  \item \( P \) is the pressure of the gas (in atmospheres, atm)
  \item \( V \) is the volume of the gas (in liters, L)
  \item \( T \) is the temperature of the gas (in kelvin, K)
  \item \( N \) is the number of moles of the gas (in moles, mol)
  \item \( a \) and \( b \) are van der Waals constants specific to each gas (\(a = \text{[L}^2 \, \text{atm/mol}^2\), \(b = \text{[L/mol]}\))
  \item \( R \) is the gas constant (\(0.0821 \, \frac{\text{L} \cdot \text{atm}}{\text{mol} \cdot \text{K}}\))
\end{itemize}
The term $\frac{aN^2}{V^2}$ corrects for the attractive forces between gas molecules, while the term $Nb$ corrects for the volume occupied by the gas molecules themselves. It's important to note that the van der Waals constants $a$ and $b$ vary depending on the specific gas and are experimentally determined.

In table \ref{tab:gas_constants} below, we have listed these constants for a selection of common gases that we have used in this study.
\begin{table}[htbp]
\centering
\begin{tabular}{|c|c|c|}
\hline
Gas & $a$ [L$^2$ atm/mol$^2$] & $b$ [L/mol] \\
\hline
Ideal Gas & 0 & 0 \\
Hydrogen (H$_2$) & 0.244 & 0.0266 \\
Helium (He) & 0.0346 & 0.0237 \\
Neon (Ne) & 0.211 & 0.0174 \\
Argon (Ar) & 1.355 & 0.032 \\
Xenon (Xe) & 4.00 & 0.051 \\
Nitrogen (N$_2$) & 1.390 & 0.0391 \\
Oxygen (O$_2$) & 1.360 & 0.0318 \\
Carbon Dioxide (CO$_2$) & 3.610 & 0.0427 \\
Methane (CH$_4$) & 2.250 & 0.0428 \\
\hline
\end{tabular}
\caption{Van der Waals constants for gases used in this study. Adapted from standard references on thermodynamics or physical chemistry.}
\label{tab:gas_constants}
\end{table}
For our data generation process, we first generate the variables for Temperature ($T$), Volume ($V$), and Moles ($N$) from Gaussian distributions with defined mean ($\mu$) and variance ($\sigma^2$). Specifically, we generate $T$, $V$, and $N$ as follows:

\begin{align}
T & \sim \mathcal{N}(\mu_T, \sigma_T^2) \\
V & \sim \mathcal{N}(\mu_V, \sigma_V^2) \\
N & \sim \mathcal{N}(\mu_N, \sigma_N^2)
\end{align}
We can then plug the gas constants and parameter distributions into the van der Waals equation to calculate the corresponding Pressure ($P$), which serves as our target variable for the synthetic dataset. 
We define the parameters for generating synthetic datasets as indicated in the tables below for experiments 1 and 2, respectively: 

\begin{table}[htbp]
\centering
\begin{tabular}{|c|c|c|}
\hline
Parameter & Mean & Std \\
\hline
Temperature ($T$) &  $\mu_T = 300 \, \text{K}$ & $ \sigma_T = 25 \, \text{K} \, $ \\
Volume ($V$) &  $\mu_V = 50 \, \text{L}$ & $ \sigma_V = 5 \, \text{L} \, $ \\
Moles ($N$) &  $\mu_N = 15 \, \text{mol}$ & $ \sigma_N = 1 \, \text{mol} \, $ \\
\hline
\end{tabular}
\caption{Parameters used for generating data for experiment 1}
\label{tab:parameters_exp1}
\end{table}

\begin{table}[htbp]
\centering
\begin{tabular}{|c|c|c|}
\hline
\multicolumn{3}{|c|}{Dataset 1:} \\
\hline
Temperature  & $\mu_{T_1} = 273 \, \text{K}$ & $\sigma_{T_1} = 50 \, \text{K}$ \\
Volume  & $\mu_{V_1} = 10 \, \text{L}$ & $\sigma_{V_1} = 1 \, \text{L}$ \\
Moles  & $\mu_{N_1} = 10 \, \text{mol}$ & $\sigma_{N_1} = 1 \, \text{mol}$ \\
\hline
\multicolumn{3}{|c|}{Dataset 2:} \\
\hline
Temperature  & $\mu_{T_2} = 300 \, \text{K}$ & $\sigma_{T_2} = 50 \, \text{K}$ \\
Volume  & $\mu_{V_2} = 9 \, \text{L}$ & $\sigma_{V_2} = 1.5 \, \text{L}$ \\
Moles  & $\mu_{N_2} = 11 \, \text{mol}$ & $\sigma_{N_2} = 1 \, \text{mol}$ \\
\hline
\end{tabular}
\caption{Parameters used for generating data for experiment 2}
\label{tab:parameters_exp2}
\end{table}

\begin{itemize}
\item For experiment 1, we use identical distributions for the input variables (Temperature, Volume and Moles), but different van der Waals constant for the various gases. 
\item For experiment 2, we use the ideal gas approximation for both datasets (a = b = 0), but rather change the distributions for the input variables, as illustrated in table \ref{tab:parameters_exp2}

\end{itemize}

\section{Results and Discussion}
\label{sec:Results}
This section presents a summary of the key findings from the study. For comprehensive details and access to the code implementation of the analysis methodology, readers are directed to the material available on GitHub: \url{https://github.com/veflo/uncert_quant}

\subsection{Experiment 1:}

\begin{figure}[htbp]
    \centering
    \begin{subfigure}{0.49\linewidth}
        \includegraphics[width=\linewidth, height=4.3cm]{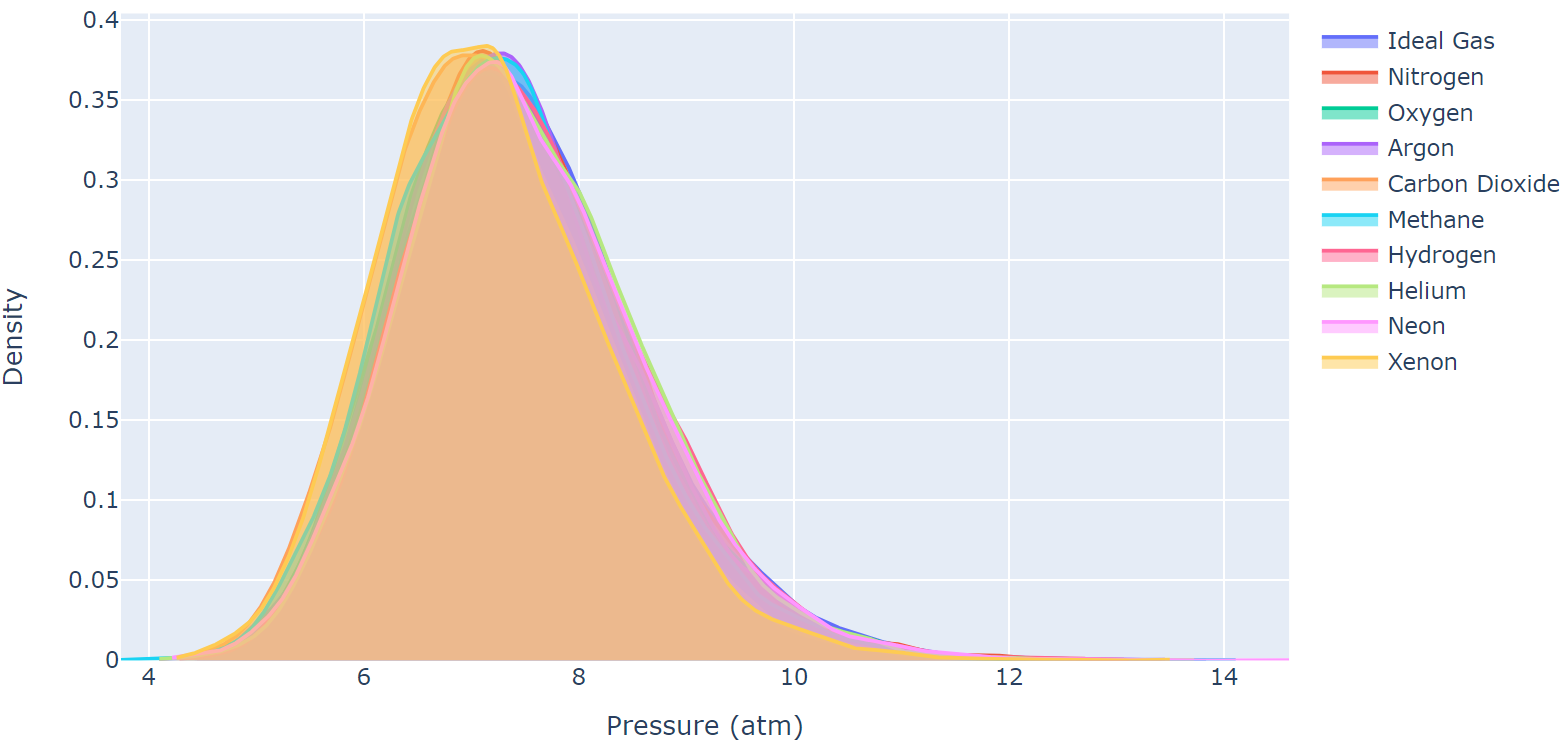}
        \caption{Pressure distribution, as calculated from van der Waals equation.}
        \label{subfig:exp1_pressure_dist}
    \end{subfigure}
    \begin{subfigure}{0.49\linewidth}
        \includegraphics[width=\linewidth]{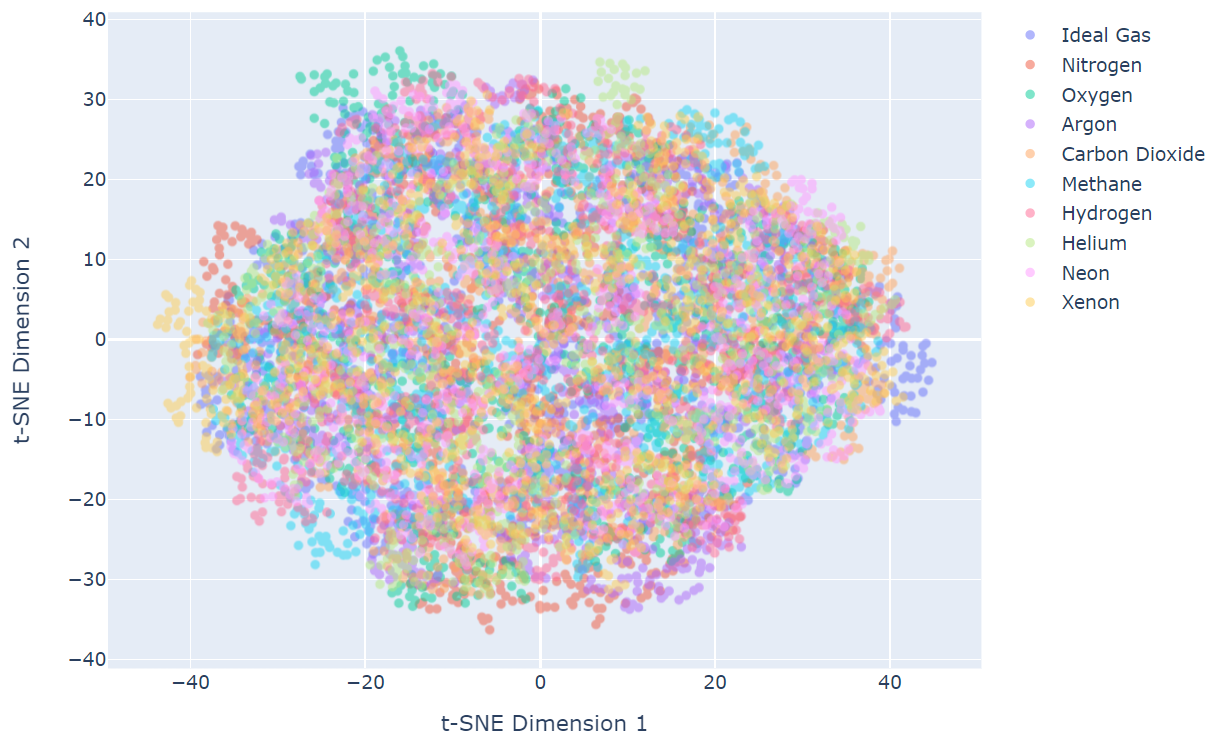}
        \caption{t-SNE visualization}
        \label{subfig:exp1_t-SNE}
    \end{subfigure}
    \caption{Pressure distribution and t-SNE plot for the various gases}
    \label{fig:exp1}
\end{figure}

We begin by generating a synthetic dataset for the selection of gases, as described in further detail in Section \ref{sec:data_generation_van_der_waals}. The variables Temperature, Moles, and Volume are uniformly generated for all gases, while the corresponding pressure is calculated using the van der Waals equation. This equation introduces corrections to the ideal gas approximation, resulting in varying degrees of adjustments based on the properties of each gas. Figure \ref{subfig:exp1_pressure_dist} illustrates the resulting pressure distribution, highlighting subtle differences among the different gases. In addition to the generated datasets for the other gases, we have also set aside a small subset of the ideal gas data as an "in distribution" test set. 

To visualize the degree of similarity between the datasets, we employ t-SNE, a popular technique for visualizing high-dimensional data \cite{maaten2008visualizing}. The t-SNE plot, as shown in Figure \ref{subfig:exp1_t-SNE}, reveals no clear clusters or outliers in this case. However, given the subtle differences between the gases and that the ideal gas approximation is a fairly accurate description in many scenarios, this result is not unexpected. 
To perform a more quantitative investigation of data similarity and distribution shift beyond visual comparison, we utilize KL-divergence and Jensen Shannon Distance, as introduced in Section \ref{sec:data_similarity}. 

\begin{figure}[htbp]
    \centering
    \includegraphics[width=.7\linewidth]{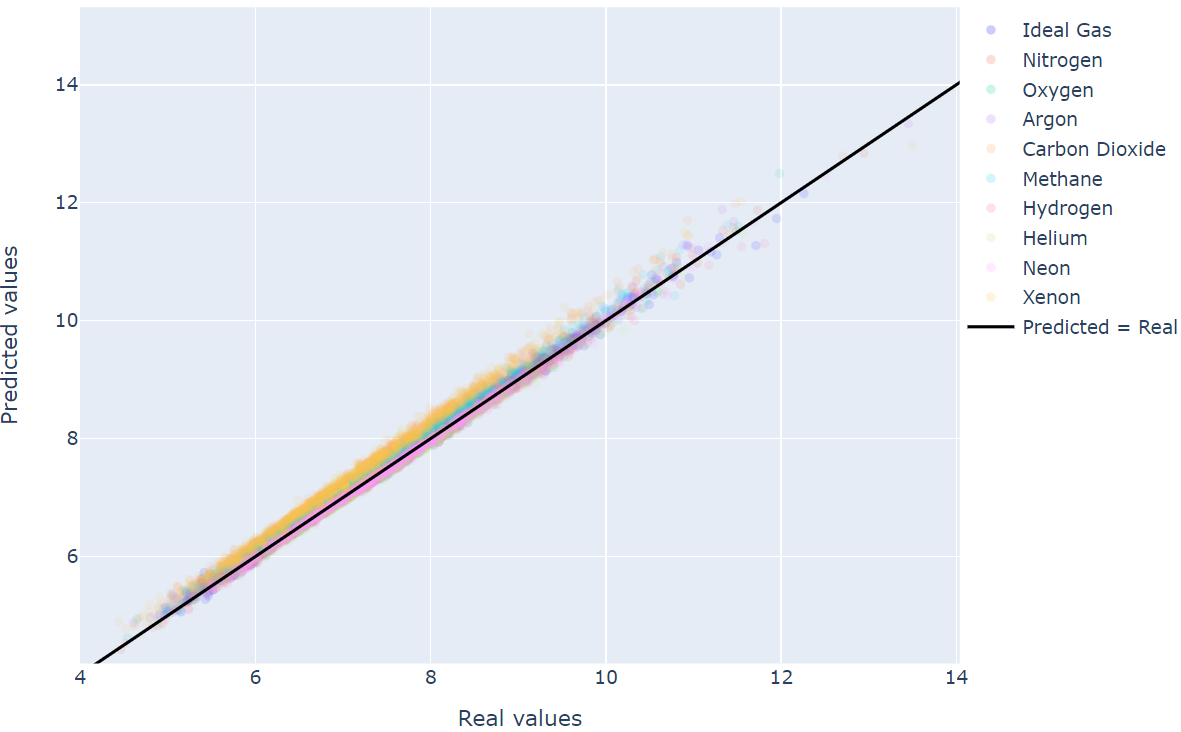}
    \caption{Predicted vs Real Pressure, Experiment 1}
    \label{fig:exp1_pred_vs_real}
\end{figure}

Next, we train our machine learning model using the ideal gas training data, and then evaluate its generalization to the other datasets. Figure \ref{fig:exp1_pred_vs_real} illustrates the model's predictions vs. actual pressure values for the different gases.
In addition to the visual comparison of the predictions, we also compute the Mean Absolute Percentage Error (MAPE). In Table \ref{tab:exp1_results}, we summarize both the KL-divergence, JS Distance, and MAPE between the ideal gas training dataset and other gases.

\begin{table}[htbp]
\centering
\begin{tabular}{|l|c|c|c|}
\hline
Dataset & KL-Div & JS Distance & MAPE \\
\hline
Ideal Gas (subset) & 0.10 & 0.14 & 0.48\\
Nitrogen & 0.33 & 0.26 & 0.73\\
Oxygen & 0.57 & 0.33 & 0.86\\
Argon & 0.52 & 0.32 & 0.84\\
Carbon Dioxide & 5.49 & 0.73 & 3.30\\
Methane & 1.75 & 0.49 & 1.56\\
Hydrogen & 0.31 & 0.25 & 0.64\\
Helium & 0.55 & 0.33 & 0.76\\
Neon & 0.14 & 0.17 & 0.54\\
Xenon & 6.11 & 0.75 & 3.55\\
\hline
\end{tabular}
\caption{KL-div, JS Distance, and prediction error between the training dataset (Ideal Gas) and the datasets for the other gases}
\label{tab:exp1_results}
\end{table}

From these results, we note that smaller corrections to the van der Waals equation typically yield lower KL-divergence and JS Distance, indicating greater similarity to the training dataset. Consequently, datasets more closely resembling the ideal gas approximation result in lower prediction errors. 
Furthermore, it is worth noting that despite being drawn from the same distribution, we observe a non-zero KL-divergence and JS distance between the training data and the "in-distribution" test set, referred to as "Ideal Gas (subset)." This observation underscores the inherent statistical variations resulting from a finite number of generated data points, which in this case were limited to 10,000 samples.

This phenomenon reflects a conventional practice in assessing machine learning models, where a portion of the training data is reserved for testing purposes. However, when the test data originates from the same distribution as the training data, it often leads to overly optimistic estimates of model accuracy. These discrepancies often become apparent upon deployment "in the wild," where the model may encounter a different data distribution, resulting in a corresponding degradation in accuracy.

\begin{figure}[htbp]
    \centering
    \begin{subfigure}{0.49\linewidth}
        \includegraphics[width=\linewidth]{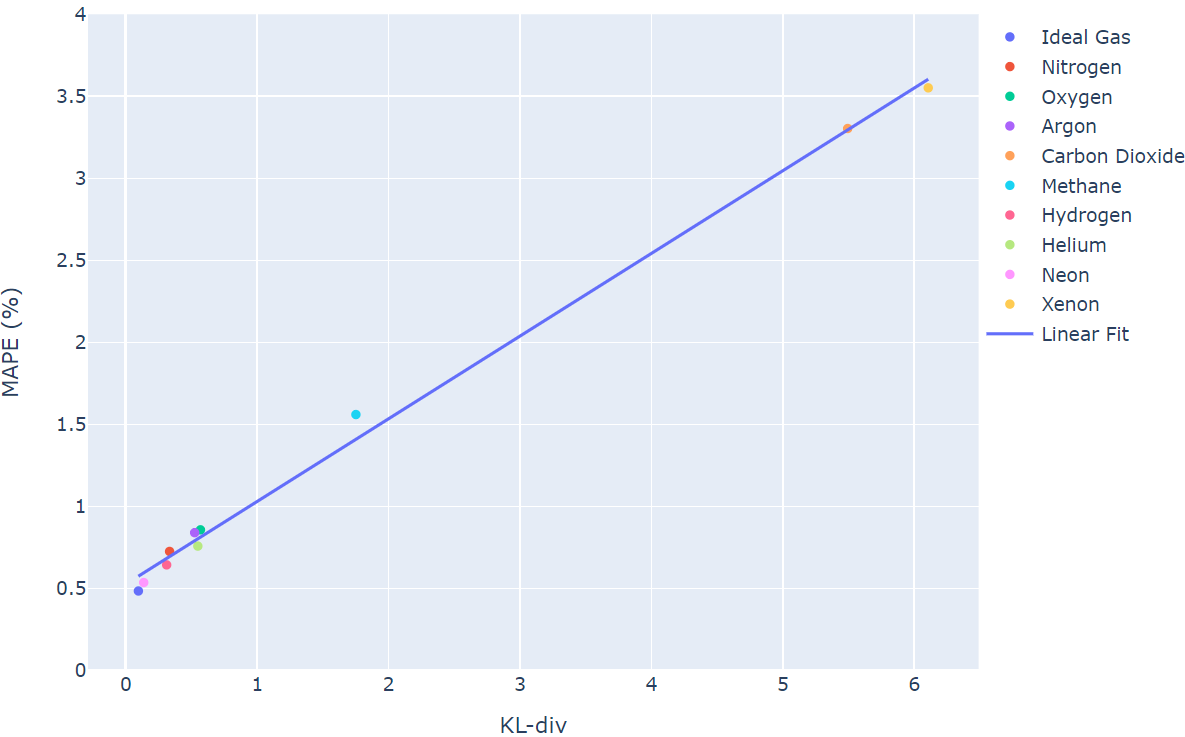}
        \caption{KL-Divergence vs. Prediction error}
        \label{subfig:KL_vs_error}
    \end{subfigure}
    \begin{subfigure}{0.49\linewidth}
        \includegraphics[width=\linewidth]{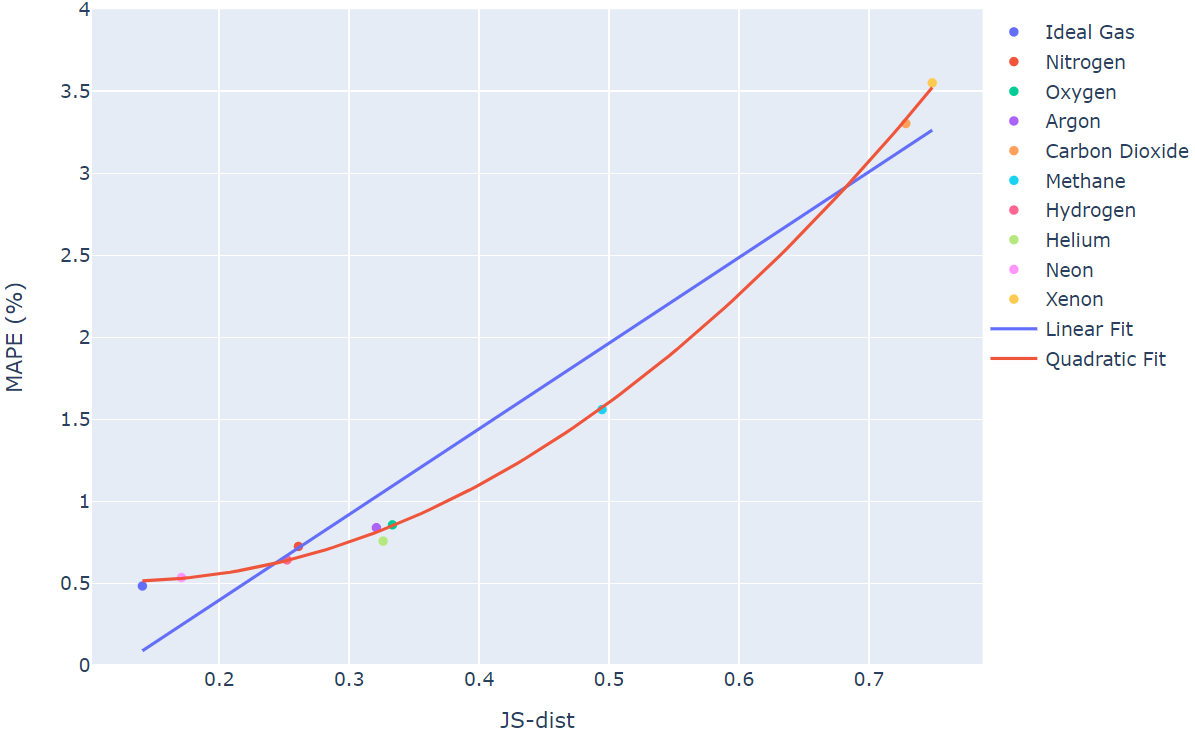}
        \caption{JS-Distance vs. Prediction error}
        \label{subfig:JS_vs_error}
    \end{subfigure}
    \caption{Comparison of KL-Divergence and JS-Distance vs. Prediction Error, demonstrating a clear correlation between the degree of distribution shift and the consequent decline in model accuracy.}
    \label{fig:KL_JS_MAPE}
\end{figure}

As illustrated in Figure \ref{subfig:KL_vs_error}, we observe a linear trend between KL-divergence and MAPE, while the relationship with JS-Distance in Figure \ref{subfig:JS_vs_error} appears less pronounced. This discrepancy may result from JS-Distance being a bounded metric between [0,1], unlike the unbounded nature of the KL-divergence. However, a deeper understanding of this scaling behavior would require more detailed investigations, which are beyond the scope of this study.

In summary, the outcomes of Experiment 1 underscore the potential use of KL-divergence and JS Distance as quantitative indicators of distribution shift, serving as predictors of reduced model accuracy and increased uncertainty. These metrics thus offer valuable insights for monitoring the performance and robustness of machine learning models in real-world scenarios, where varying degrees of distribution shift are frequently encountered.

\subsection{Experiment 2}
Experiment 2 extends the analysis of Experiment 1 by primarily focusing on quantifying prediction uncertainty, thereby enhancing the assessment of model reliability. While Experiment 1 simulated "target drift", where the correlations between the input variables and the target differ across various datasets, Experiment 2 focuses on "covariate shift". Here, the distribution of the input variables themselves has changed, while the correlations to the target variable remain constant

\begin{figure}[htbp]
    \centering
    \begin{subfigure}{0.49\linewidth}
        \includegraphics[width=\linewidth]{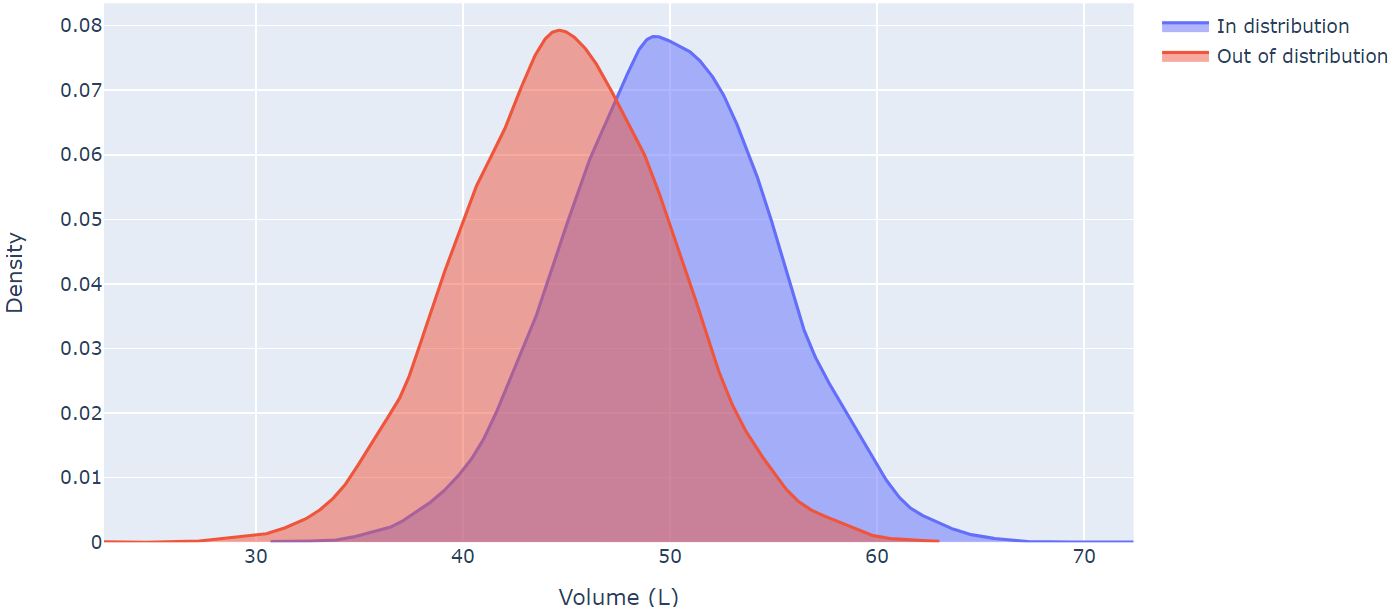}
        \caption{Volume}
        \label{subfig:exp2_volume}
    \end{subfigure}
    \begin{subfigure}{0.49\linewidth}
        \includegraphics[width=\linewidth]{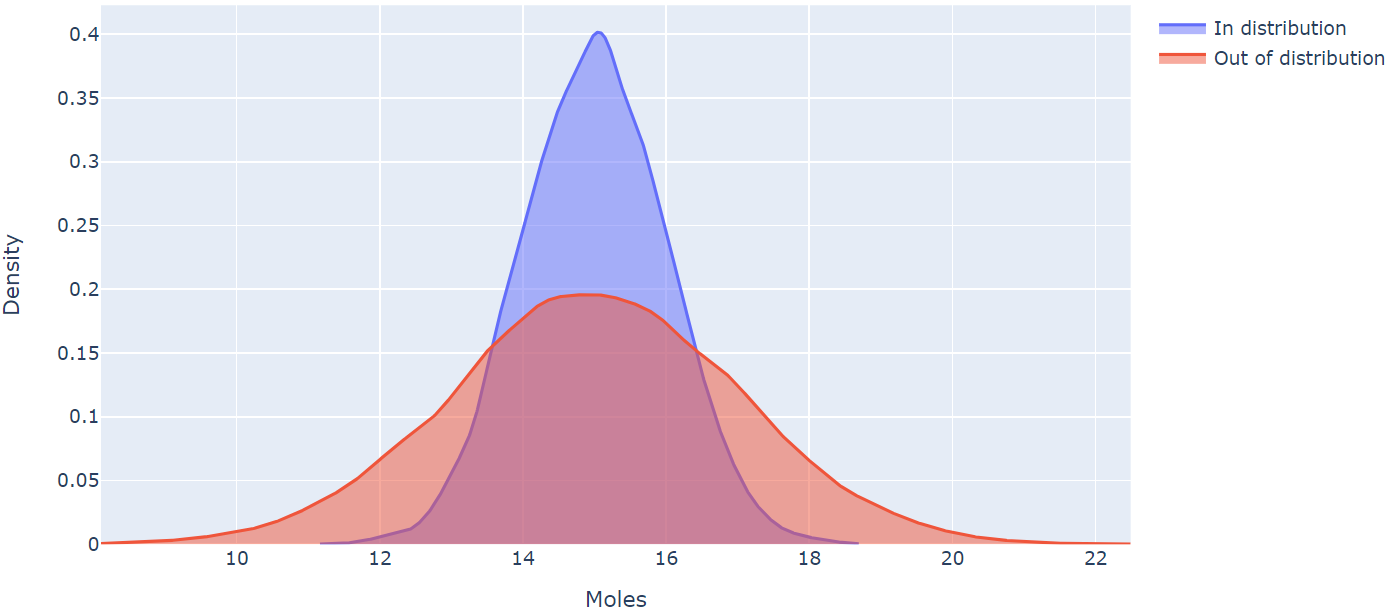}
        \caption{Moles}
        \label{subfig:exp2_moles}
    \end{subfigure}
    \begin{subfigure}{0.49\linewidth}
        \includegraphics[width=\linewidth]{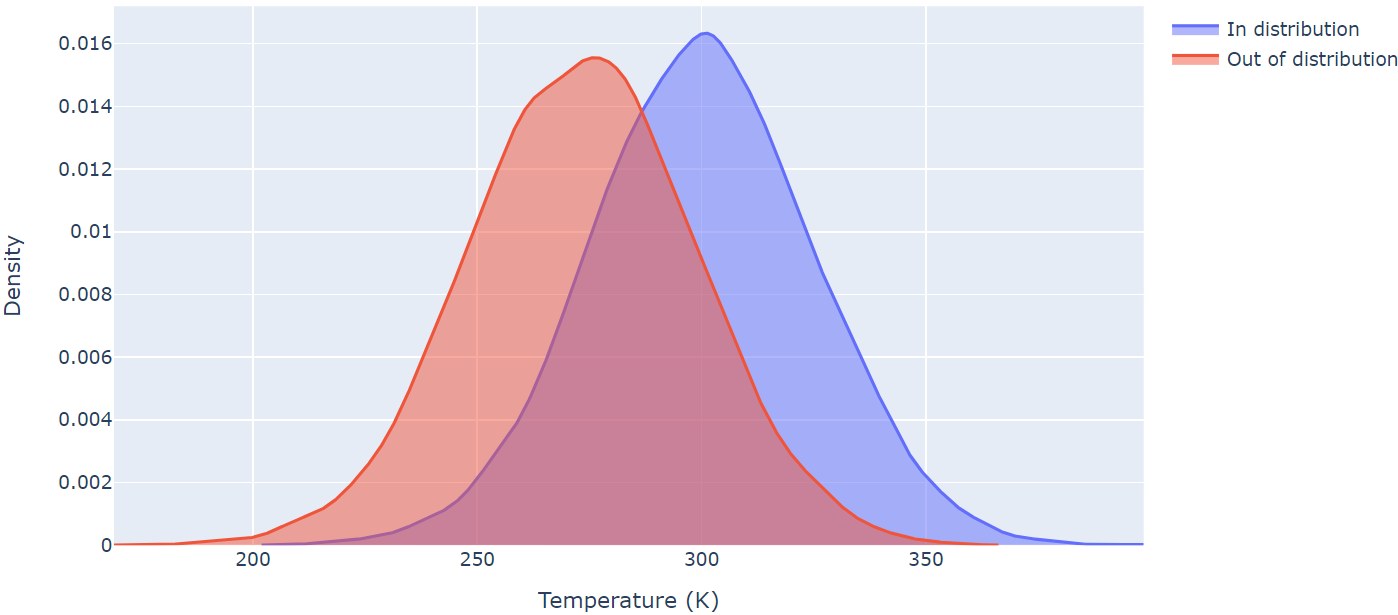}
        \caption{Temperature}
        \label{subfig:exp2_temp}
    \end{subfigure}
    \begin{subfigure}{0.49\linewidth}
        \includegraphics[width=\linewidth]{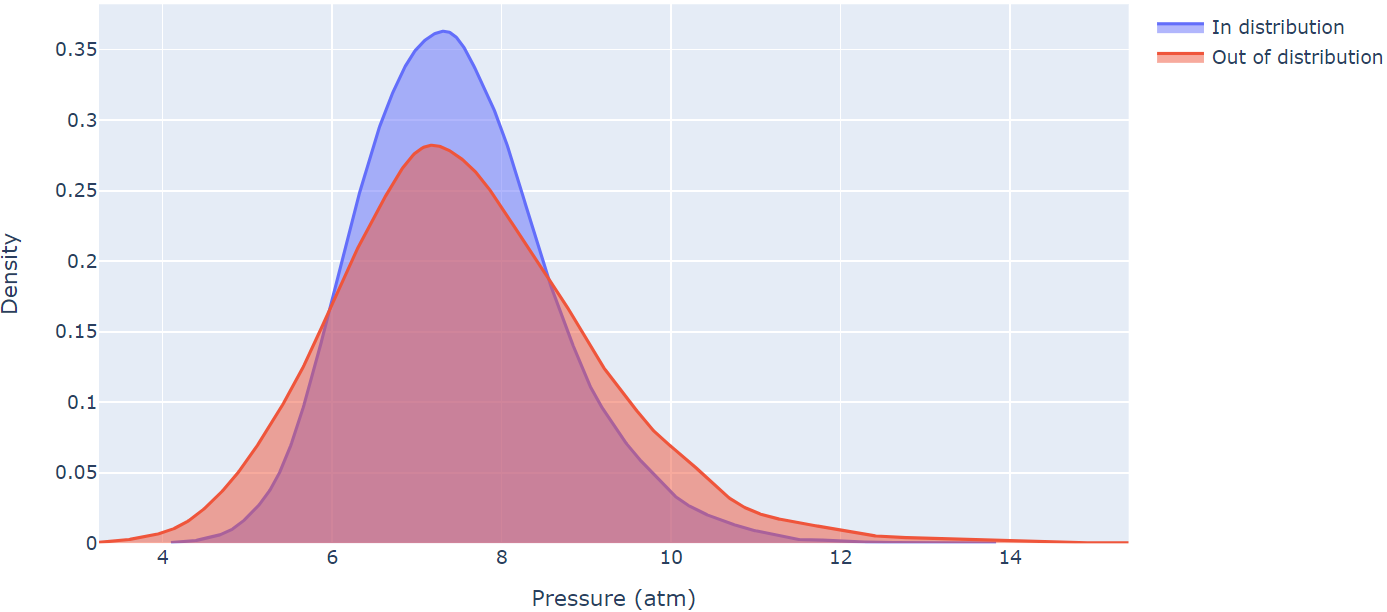}
        \caption{Pressure}
        \label{subfig:exp2_pressure}
    \end{subfigure}
    \caption{Distribution of Volume, Moles, Temperature, and the resulting calculated Pressure for the train ("in distribution") vs. test ("out-of-distribution") datasets}
    \label{fig:exp2_distributions}
\end{figure}

Specifically, we generate data using the ideal gas approximation for two datasets, with the differences being determined by the parameters of the distributions used for data generation. The resulting data distributions for the generated data are illustrated in Figure \ref{fig:exp2_distributions}. Further details on the data generation process are also described in Table \ref{tab:parameters_exp2} in section \ref{sec:data_generation_van_der_waals}.

We also here calculate the corresponding KL-divergence and JS-Distance between the "in distribution" training data vs. the "out-of-distribution" test set, as presented in Table \ref{tab:exp2_KL_JS_MAPE}. This is done by setting aside a subset of the training data as an "in distribution" test set, following a similar procedure as in Experiment 1. 
\begin{table}[htbp]
\centering
\begin{tabular}{|l|c|c|c|}
\hline
Dataset & KL-Div & JS Distance & MAPE \\
\hline
In distribution (subset) & 0.26 & 0.16 & 0.26\\
Out of distribution & 1.20 & 0.47 & 0.60\\
\hline
\end{tabular}
\caption{KL-div, JS Distance, and prediction error for the "in distribution" vs "out-of-distribution" datasets}
\label{tab:exp2_KL_JS_MAPE}
\end{table}

\begin{figure}[htbp]
\centering
\includegraphics[width=.7\linewidth]{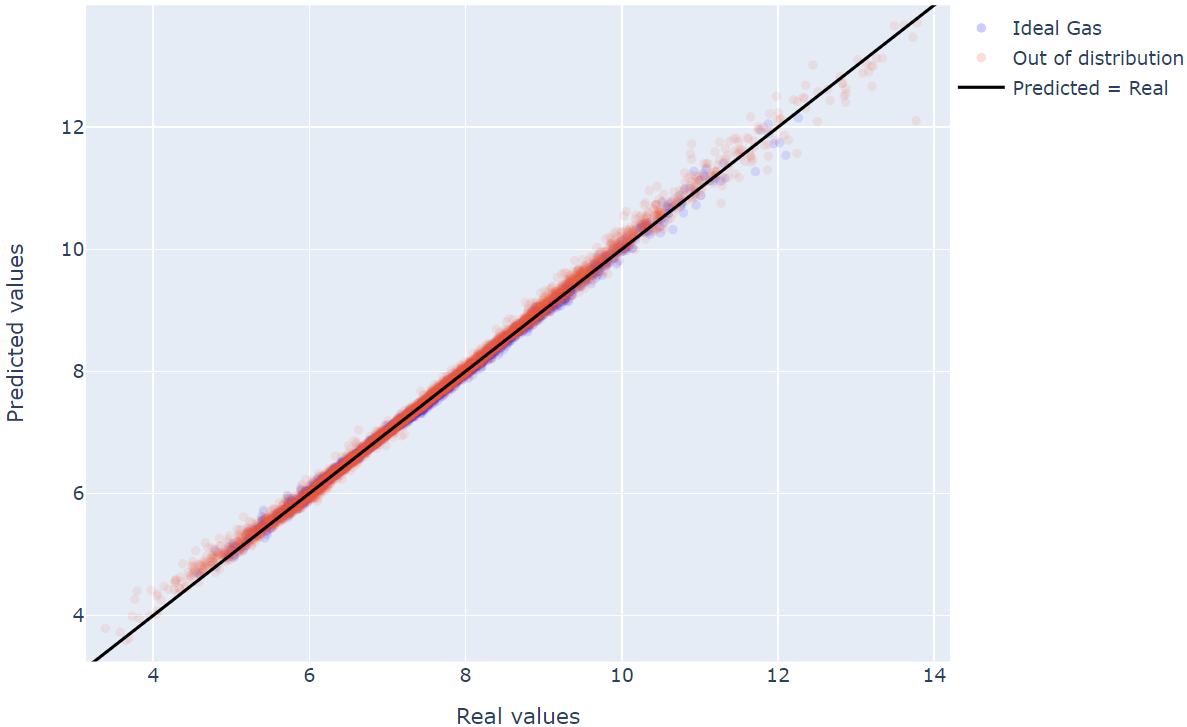}
\caption{\label{fig:exp2_pred_vs_real}Predicted vs. Real Pressure, Experiment 2 }
\end{figure}

We then proceed to train our machine learning model using the training data and utilize it to make predictions for both the "in-distribution" and "out-of-distribution" test sets.  
Figure \ref{fig:exp2_pred_vs_real} illustrates the predicted vs. real pressure values. The largest deviations occur in regions of low and high pressure, where the model encounters data outside the training distribution, as shown in Figure \ref{subfig:exp2_pressure}. The associated prediction error (MAPE) is also provided in Table \ref{tab:exp2_KL_JS_MAPE}, for comparison between the "in-distribution" and "out-of-distribution" test sets.

\subsubsection{Uncertainty Quantification}
To evaluate model uncertainty, we employ the Monte Carlo Dropout technique, as discussed in detail in section \ref{sec:monte_carlo_dropout}. This technique involves conducting N=100 forward passes through the network for each input, resulting in multiple predictions $y_i$ sampled from the model's predictive distribution. By calculating the mean and standard deviation of these predictions for each input data point, we thus obtain an estimate of the model's epistemic uncertainty.

\begin{figure}[htbp]
\centering
\includegraphics[width=.7\linewidth]{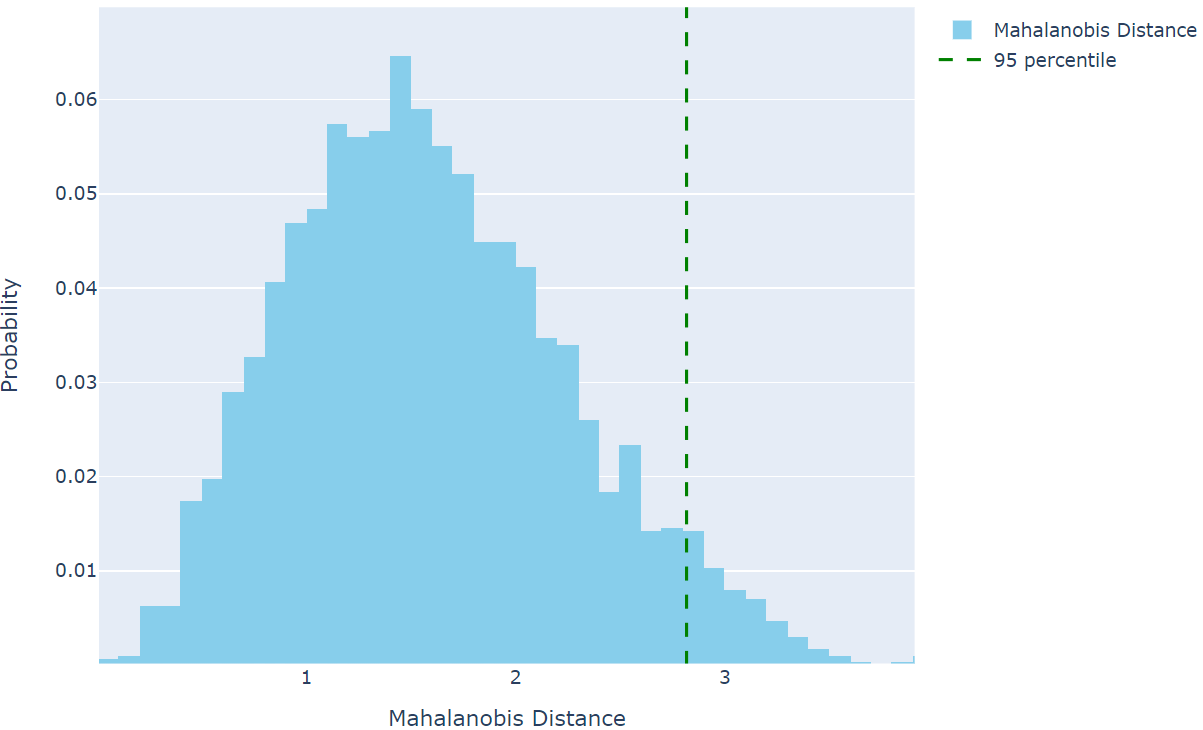}
\caption{\label{fig:mahalanobis_dist}Mahalanobis distance for the "In distribution" dataset }
\end{figure}

We then compute the Mahalanobis distance, as introduced in section \ref{sec:mahalanobis}, which serves as a metric to assess how far each datapoint deviates from the training data distribution. Figure \ref{fig:mahalanobis_dist} illustrates the distribution of Mahalanobis distances for the training data, including the limit indicating the 95th percentile of this distribution. 
By computing the Mahalanobis distance for all datapoints in the test set, we can then gain insights into the error and uncertainty of model predictions relative to their distance from the training distribution. 

\begin{figure}[htbp]
    \centering
    \begin{subfigure}{0.49\linewidth}
        \includegraphics[width=\linewidth]{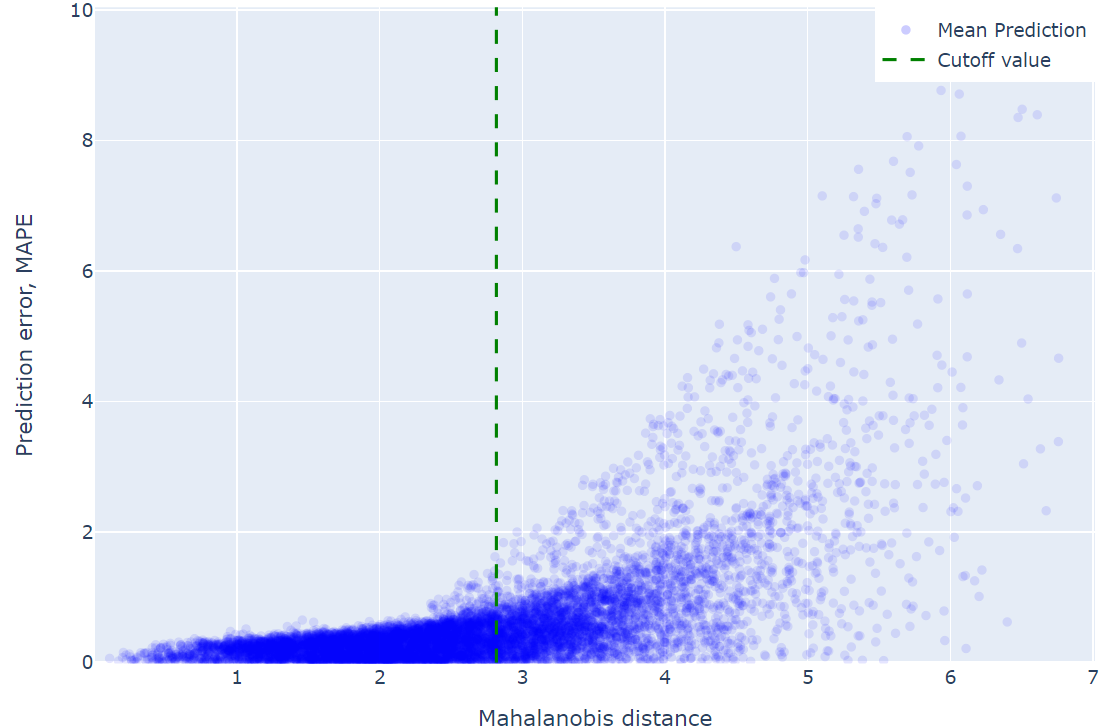}
        \caption{Mahalanobis distance vs. prediction error}
        \label{subfig:Mahalanobis_vs_error}
    \end{subfigure}
    \begin{subfigure}{0.49\linewidth}
        \includegraphics[width=\linewidth]{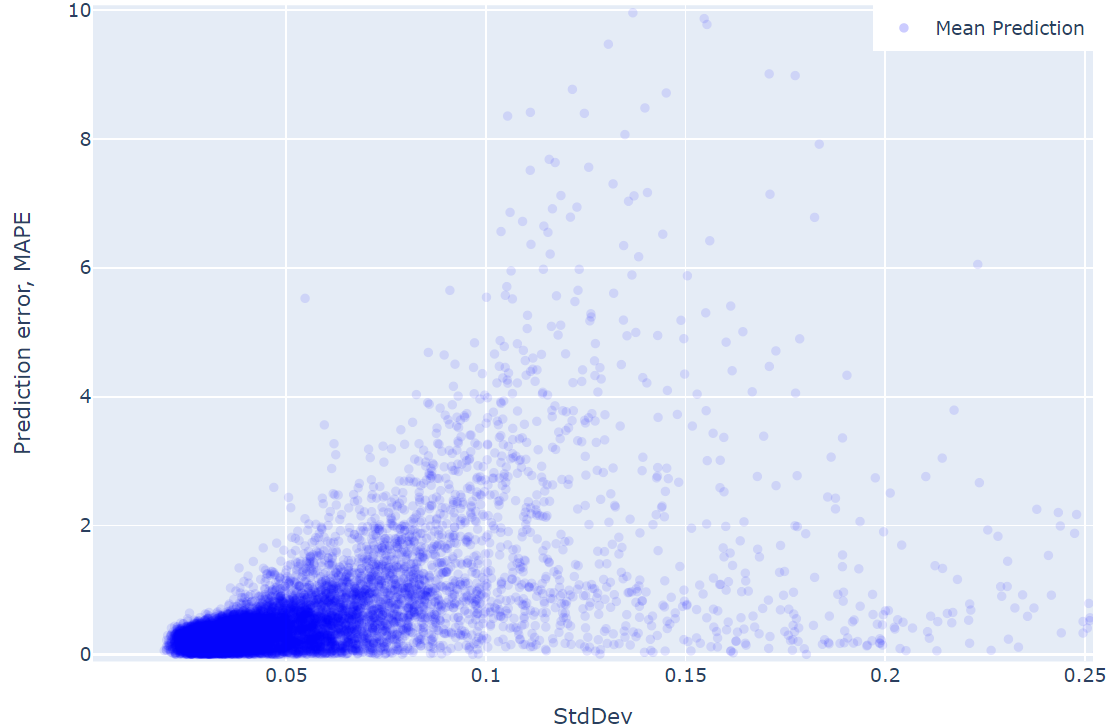}
        \caption{Standard deviation vs. prediction error}
        \label{subfig:std_vs_error}
    \end{subfigure}
    \caption{Comparison of Mahalanobis distance and standard deviation as indicators of prediction uncertainty.}
    \label{fig:comparison}
\end{figure}

As illustrated in Figure \ref{subfig:Mahalanobis_vs_error}, datapoints with a low Mahalanobis distance, indicating proximity to the training distribution, generally exhibit lower prediction errors. Conversely, an increase in the Mahalanobis distance, particularly beyond the 95th percentile cutoff value, signifies a departure from the training distribution. This correlates with a corresponding rise in prediction errors and the spread of predictions, indicating increased uncertainty. 

This behavior is further explored in Figure \ref{subfig:std_vs_error}, which illustrates the relationship between prediction error and the standard deviation of predictions, commonly used as an indicator of model uncertainty in the Monte Carlo Dropout technique. 
While low standard deviation values often coincide with low prediction errors, the correlation is less pronounced compared to the Mahalanobis distance.

These findings suggests that the Mahalanobis distance may offer superior, or at least complementary, insights into prediction uncertainty and error compared to relying solely on the Monte Carlo Dropout technique. Moreover, the Mahalanobis approach offers significant computational advantages over Monte Carlo methods. By requiring only a single forward pass per prediction instead of multiple iterations, it greatly enhances the practicality and efficiency of the approach.

Importantly, the Mahalanobis distance serves as a metric for assessing prediction reliability on a per-data-point basis. Data points with Mahalanobis distances well below the cutoff can be considered relatively accurate and trustworthy, while those exceeding the cutoff may require further investigation or precautionary measures.

Combining the Mahalanobis approach with Monte Carlo Dropout techniques, or other methods for uncertainty quantification, holds promise for a more comprehensive assessment of model uncertainty and determining when model predictions can be trusted or not. This allows for making more informed decisions based on the trustworthiness of the predictions, thereby enhancing the robustness of machine learning systems deployed in dynamic real-world environments.

\section{Summary and Conclusions}
\label{sec:summary}
This paper presents two experiments aimed at investigating the impact of distribution shift on machine learning model performance and uncertainty quantification.

In Experiment 1, synthetic datasets representing different gases were generated, and a machine learning model was trained using the ideal gas dataset. By quantifying the degree of distribution shift through the KL-divergence and Jensen Shannon Distance between the various datasets, we observed a clear correlation between distribution shift and prediction accuracy. Higher errors were evident for datasets deviating further from the ideal gas distribution, emphasizing the critical role of monitoring distribution shift for maintaining model performance in real-world applications. Furthermore, we illustrated how KL-divergence and JS Distance serve as valuable indicators of distribution shift and predictors of model prediction accuracy degradation.

Experiment 2 focused on uncertainty quantification, particularly in scenarios of covariate shift. By employing the Monte Carlo Dropout technique for estimating model uncertainty, and by calculating the Mahalanobis distance, we correlated uncertainty with the distance of test data points from the training distribution. Our findings revealed that as data points move away from the training distribution, both prediction error and uncertainty tend to increase. The Mahalanobis distance emerged as a promising metric for assessing prediction reliability on a per-data-point basis, offering additional insights into when model predictions can be trusted or not.

While uncertainty quantification methods like Bayesian methods \cite{gawlikowski2023survey} and conformal prediction \cite{conformal_prediction} offer alternative approaches, each presents its own set of challenges and trade-offs. Bayesian methods provide a robust framework for uncertainty estimation but face challenges such as computational complexity and scalability limitations. Conversely, conformal prediction offers a pragmatic alternative but can be susceptible to shifts in data distribution over time, impacting prediction reliability.

The choice of methodology ultimately depends on the specific requirements of the application and the inherent trade-offs between computational complexity, interpretability, and performance. Rather than aiming to replace existing methodologies, this study seeks to explore complementary approaches. 
The findings presented here serve as a guide towards identifying appropriate metrics and techniques for effective and robust monitoring and governance of machine learning applications in real-world scenarios. By understanding the strengths and limitations of different uncertainty quantification methods, practitioners can make more informed decisions and tailor their approach to suit the demands of their specific application domains.

\bibliographystyle{plain} % Specify the bibliography style
\bibliography{references}

\end{document}